\documentclass{article}
\usepackage{iclr2027_conference,times}

\usepackage{amsmath,amsfonts,bm}

\def\eqref#1{equation~\ref{#1}}

\def\1{\bm{1}}

\DeclareMathAlphabet{\mathsfit}{\encodingdefault}{\sfdefault}{m}{sl}
\SetMathAlphabet{\mathsfit}{bold}{\encodingdefault}{\sfdefault}{bx}{n}

\usepackage{hyperref}
\hypersetup{hidelinks}
\usepackage{url}
\usepackage{graphicx}
\usepackage{booktabs}
\usepackage{multirow}
\usepackage{amssymb}
\usepackage{amsmath}
\usepackage{pifont}
\usepackage{xcolor}
\usepackage{float}
\usepackage{caption}
\title{PARSEE-VAD: Efficient Training-Free Online Video Anomaly Detection via Proposition-Aware Reasoning and Streaming Evidence Escalation}

\author{
Ji Wang \\
Global Institute of Future Technology\\
Shanghai Jiao Tong University\\
Shanghai, 200240, China \\
\texttt{touhouariake@sjtu.edu.cn} \\
\And
Shuangqing Zhang\\
School of Intelligence Science and Technology \\
Nanjing University \\
Suzhou, 215163, China \\
\texttt{zsq\_cs@foxmail.com}
\And
Guo-Sen Xie \\
School of Computer Science and Engineering \\
Nanjing University of Science and Technology \\
Nanjing, 210094, China \\
\texttt{guosen.xie@njust.edu.cn}
\And
Fang Zhao\thanks{Corresponding author.} \\
School of Intelligence Science and Technology \\
Nanjing University \\
Suzhou, 215163, China \\
\texttt{fzhao@nju.edu.cn}
}

\iclrfinalcopy

\begin{document}
\maketitle

\begin{abstract}
Training-free online video anomaly detection (VAD) with frozen multimodal language models faces two coupled challenges: extracting reliable current-window semantics under causal and computational constraints, and maintaining temporal continuity without repeatedly transmitting high-dimensional history. Encoding history through text can compress visual evidence and introduce semantic bias, whereas retaining visual history expands multimodal context. We introduce PARSEE-VAD, a two-module framework that separates semantic evidence acquisition from score-state evolution. Proposition-Aware Reasoning (PAR) extracts structured propositional evidence from the current causal window and conditionally activates more specific queries when coarse evidence warrants further refinement. By sharing a reusable causal visual prefix across queries, PAR reduces redundant computation through selective execution. Streaming Evidence Escalation (SEE) maps the acquired proposition evidence into a compact score-domain event state through current evidence escalation, then propagates only the resulting bounded state across decisions to support temporal continuity. Experiments on four benchmarks demonstrate strong training-free online performance while selective routing reduces specialist computation and score-state propagation remains sparse. These results support a current-first principle for streaming multimodal inference: resolve present semantics first, then use compact historical state only to repair residual continuity gaps.
\end{abstract}

\section{Introduction}
\begin{figure}[t]
    \centering
    \includegraphics[width=\linewidth]{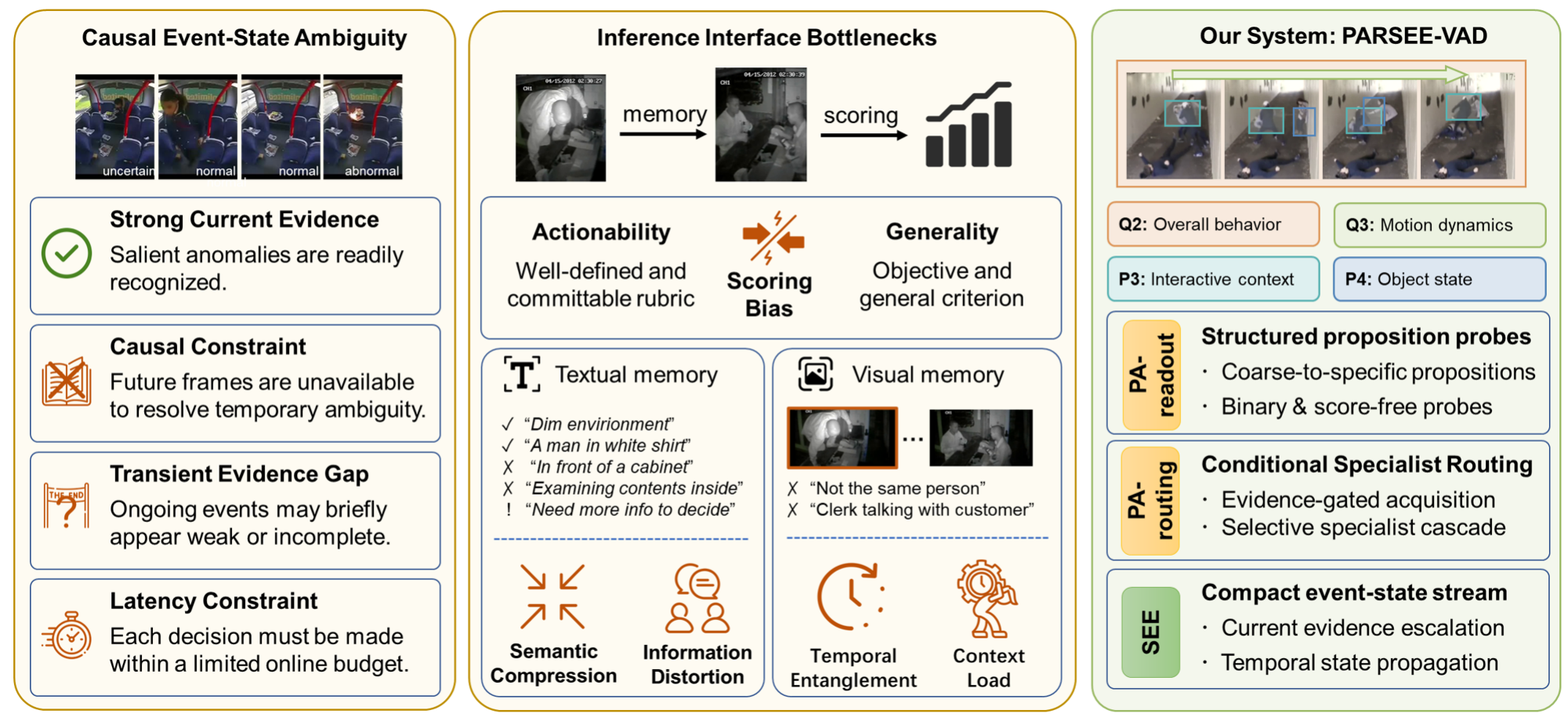}
    \caption{Motivation for PARSEE-VAD. Left: causal event-state ambiguity under no-future and online-budget constraints, including transient evidence gaps. Middle: inference-interface bottlenecks from scalar scoring and high-dimensional textual or visual memory. Right: PARSEE-VAD separates proposition-based semantic evidence acquisition from compact score-state evolution.}
    \label{fig:bottlenecks}
\end{figure}

Video anomaly detection (VAD) aims to spot rare anomalous events within long untrimmed videos. It is particularly valuable for screening massive video streams to pinpoint sparse events of interest, including accidents, violence, and other safety-critical incidents~\citep{Sultani2018,RTFM,MGFN}. Real-world practicality demands \emph{online} inference, wherein the detector produces predictions incrementally upon incoming frames without access to future information.
Building such online VAD systems conventionally requires heavy task-specific training with abundant labeled anomalies, which is costly and hard to scale across diverse real-world scenes. Frozen vision-language and multimodal large models provide a promising basis for training-free VAD because they transfer semantic priors without fitting a target-domain detector. Salient anomalies are often recognizable from the current causal observation, but an ongoing event can still produce a brief window of weak or incomplete evidence. Each online decision must also be made without future frames and within a limited inference budget. Training-free inference, however, still requires an interface that converts open-ended visual semantics into an anomaly decision. A common choice is to generate descriptions or ask the model to map the observation onto a hand-specified scalar anomaly rubric. This introduces a \emph{scoring-interface bias} (Fig.~\ref{fig:bottlenecks}): the resulting decision can depend not only on the visual evidence but also on how anomaly semantics are partitioned and described by the scoring instruction. This creates an \emph{actionability--generality tension}: detailed rubrics improve output control but inject more task-specific semantic commitments, whereas looser rubrics leave the mapping from semantics to score less constrained.

A distinct problem arises when online VAD carries evidence across time. A prevalent strategy is to propagate historical information for every new prediction. Common choices include textual memory through captions, summaries, or generated descriptions, as well as visual memory through past frames, image features, or extended visual context~\citep{LAVAD,MoniTor,Flashback,SphereVAD}. These strategies introduce distinct forms of a \emph{history transmission bottleneck} (Fig.~\ref{fig:bottlenecks}): textual memory compresses fine-grained visual evidence into language and makes the retained state dependent on generated wording, whereas visual memory expands multimodal context without explicitly separating past evidence from the current observation. These limitations raise the question of whether temporal continuity requires retransmitting high-dimensional history at all.

To address these problems, PARSEE-VAD separates semantic evidence acquisition from score-state evolution. \textbf{Proposition-Aware Reasoning (PAR)} consists of PA-readout and PA-routing. PA-readout constructs a reusable visual-prefix state and reads order-balanced signed margins from a coarse-to-specific hierarchy of binary propositions, avoiding a generated multi-level scalar anomaly score while keeping each probe tied to the same visual evidence. The hierarchy progresses from overall anomaly evidence (Q2) and visible physical dynamics (Q3) to specialist human-interaction (P3) and object/environment (P4) evidence. PA-routing then uses the available coarse evidence to acquire specialist propositions only when warranted. \textbf{Streaming Evidence Escalation (SEE)} maps the proposition evidence acquired by PAR into a compact score-domain event state, first applying current evidence corrections and then carrying only bounded score-derived state across causal decisions when the current evidence remains compatible with the preceding event. PAR therefore determines what proposition evidence is exposed and acquired, while SEE determines how that evidence changes the event state and how that state is propagated through time.
Extensive evaluations on UCF-Crime, XD-Violence, MSAD and UBnormal demonstrate competitive training-free online performance of our unified PARSEE-VAD. 

We summarize our key contributions as follows:
\begin{itemize}
    \item We propose \textbf{Proposition-Aware Reasoning (PAR)} to reduce dependence on generated scalar anomaly scoring. PAR exposes order-balanced signed proposition evidence directly and conditionally acquires more specific evidence only when warranted; a shared visual-prefix state makes these independent probes computationally efficient.
    \item We present \textbf{Streaming Evidence Escalation (SEE)}, a score-state mechanism that progressively applies current proposition evidence to the coarse Q2 state and then carries only bounded score-derived state across decisions. This unifies within-window evidence escalation with short-horizon temporal maintenance without repeatedly reintroducing high-dimensional textual or visual history.
\end{itemize}

\section{Related Work}
\subsection{Video Anomaly Detection}
Classical VAD spans supervised, weakly supervised, one-class, and unsupervised regimes~\citep{Sultani2018,RTFM,MGFN}. CLIP- and prompt-based methods add semantic modeling~\citep{CLIPTSA, STPrompt,OVVAD,HolmesVAU}; RVT~\cite{RVT2026} introduces a vision-centric reconstructive objective to supervise visual outputs under a weakly supervised paradigm. More recently, TD-VAD~\cite{TDVAD} reduces reliance on visual training data by learning from LLM-generated textual sequences. VadCLIP~\citep{VadCLIP} is a representative weakly supervised vision--language baseline, whereas PARSEE-VAD keeps the multimodal language model frozen and performs training-free online inference.

\subsection{Training-Free and MLLM-Based VAD}
Training-free VAD exploits pretrained models without fitting a target-domain detector. LAVAD~\citep{LAVAD} reasons through generated language, while VADTree~\citep{VADTree}, PANDA~\citep{PANDA}, URF-HVAA~\citep{URFHVAA}, and SphereVAD~\citep{SphereVAD} explore hierarchical, agentic, or frozen-representation inference. DR-VAD~\citep{DRVAD} combines multi-VLM descriptions with definition-guided reasoning and temporal refinement, while PRISM~\citep{PRISM} builds lightweight semantic anomaly axes over frozen multimodal embeddings. MACD~\citep{MACD} explicitly targets online VAD through conditionally invoked multi-agent counterfactual reasoning. Across this literature, current evidence can be mediated by generated language, definition-guided judgments, or semantic scoring axes before the final anomaly decision. PARSEE-VAD instead treats direct signed proposition readout as the evidence interface itself: PAR exposes coarse evidence without first generating a multi-level scalar score and conditionally extends semantic inference only when that evidence warrants specialist acquisition.

\subsection{Online VAD and Temporal Memory}
Online VAD excludes future frames, although clip- or window-level systems may commit decisions at discrete anchors. REWARD~\citep{REWARD}, MoniTor~\citep{MoniTor}, Flashback~\citep{Flashback}, online SphereVAD~\citep{SphereVAD}, and MACD~\citep{MACD} use different forms of streaming memory, reasoning, or conditional computation. Flashback, for example, operates on fixed-length video segments before refining segment-level scores into a frame-wise curve~\citep{Flashback}. PARSEE-VAD separates two roles that are often intertwined: PAR exposes and selectively acquires proposition evidence from the current causal observation, while SEE maps that evidence into bounded score-state updates and carries only this compact state across decisions. Because online methods use different native decision units and cadences, their published accuracy numbers are not a latency-normalized systems comparison.

\section{Method}
\begin{figure}[t]
    \centering
    \includegraphics[width=\linewidth]{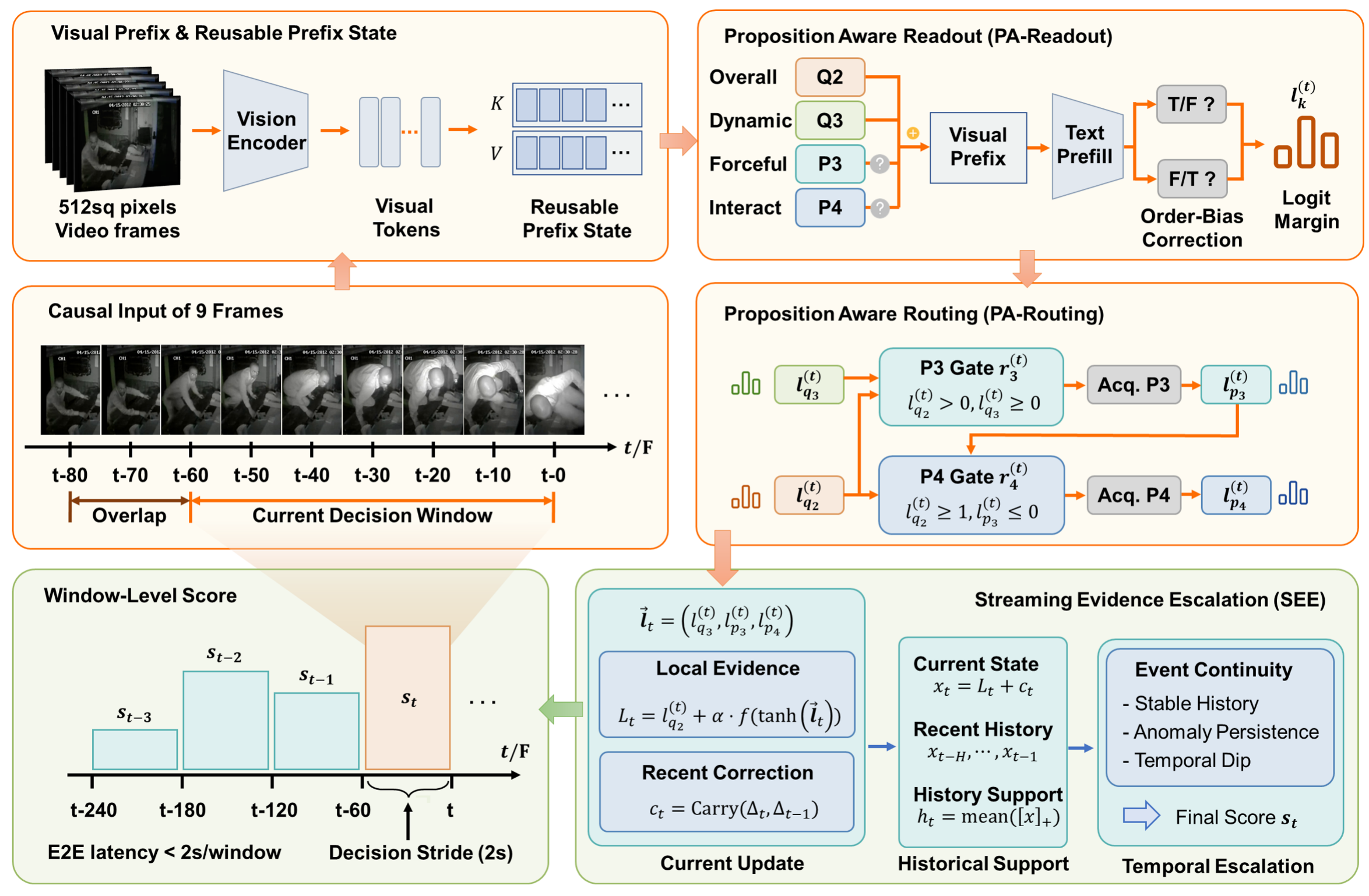}
    \caption{Overview of PARSEE-VAD. One causal visual-prefix state is reused across proposition queries. PA-readout yields order-balanced margins $\ell_k^{(t)}$, and PA-routing selectively acquires P3/P4. SEE converts the available proposition evidence into the current-evidence state $L_t$ through routed escalation $f$, then carries bounded score-derived state across decisions to emit $s_t$. The stream uses a 60-frame (2-s) decision stride.}
    \label{fig:framework}
\end{figure}

At decision anchor $t_k$, let $X_{t_k}\subseteq V_{\le t_k}$ denote the causal observation formed from frames available up to $t_k$, and let $\mathcal{H}_{k-1}$ denote the compact state carried from preceding decisions. PARSEE-VAD emits one score for the completed interval $I_k=(t_{k-1},t_k]$ through the detector $F$,
\begin{equation}
 s_k=F(X_{t_k},\mathcal{H}_{k-1}),\qquad X_{t_k}\subseteq V_{\le t_k}.
\label{eq:causal_anchor}
\end{equation}
For frame-level benchmark evaluation, $s_k$ is assigned retrospectively to every frame in $I_k$ after the interval closes; this mapping does not imply earlier score availability, which is analyzed separately in Appendix~\ref{app:diagnostics}. For notational simplicity, the remaining equations use $t$ for a generic decision anchor. Figure~\ref{fig:framework} summarizes PAR evidence acquisition followed by SEE score-state escalation within and across decisions.

\subsection{Proposition-Aware Reasoning}
PAR is designed to expose signed current-window evidence without first mapping the observation through a generated multi-level anomaly rubric, while keeping specialist probes independent of earlier semantic judgments. For each decision anchor, nine ordered frames are nominally sampled at offsets $[-80,-70,\ldots,0]$. Indices before stream onset are clamped to the first available frame (frame 0); the run's native decision anchors are otherwise preserved, including the small number of MSAD videos with a shorter regular decision stride.

\paragraph{PA-readout.}
The frozen vision encoder is evaluated once to construct a reusable visual-prefix state, from which proposition-specific text prefills branch. This separates shared visual computation from the conditional semantic continuations required by PA-routing and enables efficient independent probing: every proposition starts from the same visual evidence without inheriting the textual context of another query.

PAR defines four semantic propositions that progress from coarse to specific evidence. Q2 asks whether the current observation is abnormal overall. Q3 isolates whether directly visible physical dynamics are present. When more specific evidence is needed, P3 probes forceful human interaction and P4 probes concrete object/environment interaction. Q2 and Q3 are always read, whereas P3 and P4 are specialist propositions acquired conditionally. The proposition wording is specified from observable-event semantics: Q3 isolates directly visible change, P3 requires forceful bodily interaction, and P4 isolates consequential object, vehicle, or environmental interaction.

Each proposition is posed in both forward and reversed A/B order; we average the sign-aligned final-token margins to reduce option-position bias:
\begin{equation}
 \ell_k^{(t)}=
 \frac{(z_B^F-z_A^F)+(z_A^R-z_B^R)}{2},
 \qquad k\in\{q_2,q_3,p_3,p_4\}.
\label{eq:abmargin}
\end{equation}
The resulting $\ell_k^{(t)}$ is used directly as proposition evidence; no decoded caption or generated scalar score is required.

\paragraph{PA-routing.}
P3 is acquired for anomaly-positive dynamic windows, while P4 additionally requires stronger Q2 evidence not positively explained by P3:
\begin{equation}
\begin{aligned}
 r_3^{(t)}&=\mathbb{I}\!\left[\ell_{q_2}^{(t)}>0\;\wedge\;\ell_{q_3}^{(t)}\ge0\right],\\
 r_4^{(t)}&=\mathbb{I}\!\left[r_3^{(t)}=1\;\wedge\;\ell_{q_2}^{(t)}\ge1\;\wedge\;\ell_{p_3}^{(t)}\le0\right].
\end{aligned}
\end{equation}
Zero thresholds follow the signed-margin semantics; the stricter Q2 threshold is a prior for the narrower P4 branch. PA-routing ends with the acquired proposition margins; SEE handles the subsequent score-state update.

\subsection{Streaming Evidence Escalation}
SEE operates entirely in the score/state domain. Starting from the coarse Q2 margin, it first forms a current-evidence state from the proposition evidence exposed by PAR, then carries only bounded score-derived state across decisions. This keeps semantic acquisition tied to the current observation while temporal continuity is represented by compact state rather than repeated textual or visual memory.

\paragraph{Current evidence escalation.}
Let $[x]_+=\max(x,0)$. The routed current-window correction is
\begin{equation}
 E_t=\operatorname{clip}\!\left(
 -\tanh\!\left([-\ell_{q_3}^{(t)}]_+\right)
 +\!\sum_{k\in\{3,4\}}\!
 r_k^{(t)}
 \tanh\!\left(\ell_{q_3}^{(t)}\right)
 \tanh\!\left(\ell_{p_k}^{(t)}\right),
 -1,1
 \right),
\end{equation}
and the current-evidence state is
\begin{equation}
 \boxed{
 L_t=\ell_{q_2}^{(t)}
 +\alpha[\ell_{q_2}^{(t)}]_+E_t
 },\qquad \alpha=0.75.
\label{eq:fusion}
\end{equation}
Negative Q3 evidence suppresses Q2-positive interpretations without visible dynamics, while routed signed specialist evidence can reinforce or oppose them. Clipping bounds the correction, and $[\ell_{q_2}^{(t)}]_+$ prevents specialist evidence from creating an anomaly from a non-positive Q2 base. Figure~\ref{fig:framework} abbreviates the complete routed correction $[\ell_{q_2}^{(t)}]_+E_t$ as $f(\tanh(\vec{\ell}_t))$.

\paragraph{Recent correction carry.}
Let $\Delta_t=L_t-\ell_{q_2}^{(t)}$ denote the signed current proposition correction. When $\Delta_t<0$, SEE retains a bounded part of the preceding positive correction:
\begin{equation}
\begin{aligned}
 c_t&=\mathbb{I}[\Delta_t<0]
 \min\!\left(\rho_c[\Delta_{t-1}]_+,-\Delta_t\right),\\
 x_t&=L_t+c_t,
\end{aligned}
\qquad \rho_c=0.5.
\label{eq:carry}
\end{equation}
Thus the one-step carry can at most cancel the current negative correction rather than raise the state above coarse Q2. It stores only the raw $\Delta_t$ for the next carry step; the historical branch separately maintains recent $x_t$ states.

\paragraph{Historical support and temporal escalation.}
With $H=2$, historical support is $h_t=H^{-1}\sum_{j=1}^{H}[x_{t-j}]_+$. Temporal escalation is activated only when recent state and current coarse evidence indicate continuity while $x_t$ forms a relative dip:
\begin{equation}
 g_t=\mathbb{I}\!\left[
 \bigwedge_{j=1}^{H}x_{t-j}>0\;\wedge\;
 \ell_{q_2}^{(t)}>0\;\wedge\;
 \ell_{q_2}^{(t)}\ge\eta \ell_{q_2}^{(t-1)}\;\wedge\;
 x_t<\tau h_t
 \right],
 \qquad \eta=0.5,\ \tau=0.2.
\label{eq:see_gate}
\end{equation}
The emitted score is
\begin{equation}
 s_t=g_t\tau h_t+(1-g_t)x_t.
\label{eq:positive_valley}
\end{equation}
The gate requires positive recent state, persistent coarse Q2 support, and a relative current valley. History stores $x_t$, not $s_t$, so an escalation cannot recursively support a later escalation. Sensitivity and activation statistics are reported in Appendices~\ref{app:fixedparams} and~\ref{app:diagnostics}.

\section{Experiments}
\subsection{Experimental Setup}
We evaluate frame-level AUC on UCF-Crime~\citep{Sultani2018} and UBnormal~\citep{UBnormal}, and frame-level AUC/AP on XD-Violence~\citep{XDViolence} and MSAD~\citep{MSAD2024}. UBnormal is included as an additional open-set synthetic transfer benchmark using its official 211-video test split. Unless otherwise stated, all main experiments and analyses use the canonical nine-frame setting in Sec.~3 with frozen Qwen3.5-9B and a 512sq visual budget. UCF-Crime, XD-Violence, and UBnormal use a 60-frame decision cadence; a small number of MSAD videos retain shorter regular strides. A single PAR/SEE configuration is used across datasets; complete scalar values and sensitivity analyses are reported in Appendix~\ref{app:fixedparams}.

PARSEE-VAD natively emits one score per completed decision interval. For frame-level benchmark evaluation, each score is expanded to every frame in that completed interval after closure; Appendix~\ref{app:diagnostics} separately reports release-time availability.

\subsection{Comparison with Prior Methods}

\providecommand{\vmark}{\textcolor{green!60!black}{\ding{51}}}
\providecommand{\xmark}{\ding{55}}

\begin{table*}[t]
\centering

\caption{
Comparison with existing video anomaly detection methods.
\vmark{} and \xmark{} indicate whether each method is training-free or online. Here, online indicates that the reported inference uses only observations available by each decision time. PARSEE-VAD headline metrics use completed-interval benchmark mapping; the corresponding release-time availability view is reported in Table~\ref{tab:online_availability_all}. Provenance of the UBnormal entries is documented in Appendix~\ref{app:ubnormal_provenance}.
}
\label{tab:main_comparison}

\scriptsize
\setlength{\tabcolsep}{2.45pt}
\renewcommand{\arraystretch}{1.05}

\begin{tabular}{lcc|c|cc|cc|c}
\toprule

\multirow{2}{*}{Method}
& \multirow{2}{*}{\shortstack{Training-\\free}}
& \multirow{2}{*}{Online}
& UCF-Crime
& \multicolumn{2}{c|}{XD-Violence}
& \multicolumn{2}{c|}{MSAD}
& UBnormal
\\

\cmidrule(lr){4-4}
\cmidrule(lr){5-6}
\cmidrule(lr){7-8}
\cmidrule(lr){9-9}

&
&
& AUC (\%)
& AUC (\%)
& AP (\%)
& AUC (\%)
& AP (\%)
& AUC (\%)
\\
 
\midrule

\cite{Sultani2018}
& \xmark & \xmark
& 77.92 & -- & 73.20 & -- & -- & 50.30
\\

GODS~\cite{GODS}
& \xmark & \xmark
& 70.46 & 61.56 & -- & -- & -- & --
\\

RTFM~\cite{RTFM}
& \xmark & \xmark
& 83.31 & -- & 77.81 & 86.70 & 66.30 & 64.94
\\

RTFM (online)$^\dagger$~\cite{RTFM}
& \xmark & \vmark
& 80.63 & -- & 72.60 & -- & -- & --
\\

CLIP-TSA~\cite{CLIPTSA}
& \xmark & \xmark
& 87.58 & -- & 82.19 & -- & -- & --
\\

MGFN~\cite{MGFN}
& \xmark & \xmark
& 86.98 & -- & 80.11 & 85.00 & 63.50 & --
\\

MGFN (online)$^\dagger$~\cite{MGFN}
& \xmark & \vmark
& 81.76 & -- & 73.17 & -- & -- & --
\\

REWARD$^\dagger$~\cite{REWARD}
& \xmark & \vmark
& 86.94 & -- & 77.71 & -- & -- & --
\\

VadCLIP~\cite{VadCLIP}
& \xmark & \xmark
& 88.02 & -- & 84.51 & -- & -- & --
\\

STPrompt~\cite{STPrompt}
& \xmark & \xmark
& 88.08 & -- & -- & -- & -- & 63.98
\\

OVVAD~\cite{OVVAD}
& \xmark & \xmark
& 86.40 & -- & 66.53 & -- & -- & 62.94
\\

Holmes-VAU~\cite{HolmesVAU}
& \xmark & \xmark
& 88.96 & -- & 87.68 & -- & -- & 56.77
\\

VERA~\cite{VERA}
& \vmark & \xmark
& 86.55 & 88.26 & 70.54 & -- & -- & --
\\

TD-VAD~\cite{TDVAD}
& \xmark & \xmark
& 80.82 & 89.50 & 75.83 & -- & -- & --
\\

\midrule

UR-DMU (ZS)~\cite{URDMU}
& \vmark & \xmark
& -- & -- & -- & 74.30 & 53.40 & --
\\

CLIP (ZS)~\cite{CLIP}
& \vmark & \xmark
& 53.16 & 38.21 & 17.83 & -- & -- & --
\\

LLaVA-1.5 (ZS)~\cite{LLaVA}
& \vmark & \xmark
& 72.84 & 79.62 & 50.26 & -- & -- & --
\\

LAVAD~\cite{LAVAD}
& \vmark & \xmark
& 80.28 & 85.36 & 62.01 & -- & -- & 51.06
\\

VADTree~\cite{VADTree}
& \vmark & \xmark
& 84.74 & 90.44 & 67.82 & -- & -- & 65.80
\\

PANDA~\cite{PANDA}
& \vmark & \xmark
& 84.89 & -- & 70.16 & -- & -- & 75.78
\\

URF-HVAA~\cite{URFHVAA}
& \vmark & \xmark
& 84.28 & 91.34 & 68.07 & 85.90 & 76.40 & 68.98
\\

SphereVAD (full)$^\ddagger$~\cite{SphereVAD}
& \vmark & \xmark
& 86.38 & 95.74 & 86.99 & -- & -- & 76.46
\\

PRISM~\cite{PRISM}
& \vmark & \xmark
& 86.07 & -- & 80.86 & -- & -- & --
\\

\midrule

Flashback~\cite{Flashback}
& \vmark & \vmark
& \textbf{87.29} & \underline{90.54} & 75.13 & -- & -- & --
\\

SphereVAD (online)$^\ddagger$~\cite{SphereVAD}
& \vmark & \vmark
& 78.31 & -- & \underline{81.37} & -- & -- & \textbf{71.62}
\\

Online-LAVAD~\cite{MoniTor}
& \vmark & \vmark
& 76.06 & 76.01 & 52.63 & -- & -- & --
\\

MoniTor~\cite{MoniTor}
& \vmark & \vmark
& 82.57 & 79.11 & 55.01 & -- & -- & --
\\

MACD~\cite{MACD}
& \vmark & \vmark
& 83.85 & -- & \textbf{82.68} & -- & -- & --
\\

\midrule

\textbf{PARSEE-VAD (Ours)}
& \vmark & \vmark
& \underline{85.15}
& \textbf{93.00}
& 79.02
& \textbf{90.55}
& \textbf{82.02}
& \underline{70.18}
\\

\bottomrule
\end{tabular}

\vspace{3pt}

\parbox{\textwidth}{\footnotesize
$^\dagger$ Real-time RTFM/MGFN numbers and REWARD's reported real-time operating point are from REWARD~\cite{REWARD}; REWARD uses a 6.4-s decision period. \\
$^\ddagger$ ``Full'' and ``online'' denote the two evaluation modes reported in SphereVAD~\cite{SphereVAD}. \\
All methods are shown at their published operating points; native cadences and score-availability conventions are not latency-normalized across rows. UBnormal source provenance, including secondary comparison values, is detailed in appendix~\ref{app:ubnormal_provenance}. 
}

\end{table*}

\paragraph{Detection performance.}
Within the training-free online setting, PARSEE-VAD maintains strong accuracy on the three primary real-world benchmarks and additionally transfers to UBnormal without changing the PAR/SEE configuration. On UCF-Crime, it improves over Online-LAVAD, MoniTor, MACD, and online SphereVAD, while Flashback remains higher. On XD-Violence, PARSEE-VAD improves over Flashback by 2.46 AUC and 3.89 AP points and substantially exceeds Online-LAVAD and MoniTor; its AP remains below the values reported by online SphereVAD and MACD. On MSAD, where published training-free online results are sparse, PARSEE-VAD reaches 90.55 AUC / 82.02 AP and also exceeds the reported URF-HVAA numbers, although URF-HVAA is not online. On UBnormal, PARSEE-VAD obtains 70.18 AUC with the same 512sq configuration; online SphereVAD reports 71.62, while the fixed-constant URF-HVAA setting reports 68.98 and is not online. Appendix~\ref{app:ubnormal_provenance} records the provenance of every UBnormal value shown in Table~\ref{tab:main_comparison}, including values taken from secondary comparison tables. The resulting pattern is therefore strong accuracy under the joint constraints of frozen-model and causal online inference, rather than a universal best score on every metric. Because published online methods use different decision units, cadences, and score-availability conventions, these numbers should be read as accuracy comparisons under each reported causal protocol rather than as a latency-normalized systems ranking.

\begin{table}[t]
    \centering
    \caption{Published online operating points.}
    \label{tab:online_runtime}
    \small
    \setlength{\tabcolsep}{5.0pt}
    \begin{tabular}{lccc}
        \toprule
        Method & $D$ (s) & $C$ (s) & $C/D$ \\
        \midrule
        REWARD~\citep{REWARD} & 6.4 & 0.500 & 0.078 \\
        MoniTor~\citep{MoniTor} & 0.6 & 5.900 & 9.83 \\
        Flashback~\citep{Flashback} & 1.0 & 0.713 & 0.713 \\
        SphereVAD (online)$^{\dagger}$~\citep{SphereVAD} & 0.167 & 0.067 & 0.40 \\
        MACD$^{\ddagger}$~\citep{MACD} & 1.0 & 0.520 & 0.52 \\
        \midrule
        \textbf{PARSEE-VAD (ours)} & \textbf{2.0} & \textbf{1.372} & \textbf{0.686} \\
        \bottomrule
    \end{tabular}
    \vspace{2pt}

    \begin{minipage}{0.95\linewidth}
    \footnotesize
    \raggedright
    $D$ is the native decision interval, $C$ the reported processing time per decision, and $C/D$ the processing load relative to that cadence. Hardware and timing protocols differ across methods.
    \end{minipage}
\end{table}

\paragraph{Online operating point.}
PARSEE-VAD also sustains its native online cadence: the canonical system processes a 2-s decision interval in 1.372 s on average, corresponding to $C/D=0.686$, so the measured mean processing time remains below the 2-s decision interval. This is not the lowest published processing load---SphereVAD and MACD report smaller $C/D$ values under different hardware and timing protocols---but it establishes that the deployed proposition-based inference fits within its own cadence on average. Appendix~\ref{app:low_latency} further reports a 1-s operating point obtained by reducing the causal observation budget while leaving the PAR/SEE inference rules unchanged.

\subsection{Ablation and Analysis}
\paragraph{Accuracy--compute balance.}
We use 512sq as the default operating point because it captures most of the observed MSAD AUROC gain before runtime rises sharply at the largest visual-token budget. Beyond 512sq, the marginal accuracy gain is small while the shared visual prefix becomes substantially more expensive. A likely explanation is that moderate resolution increases the visual detail available to the proposition probes, whereas further increases mainly enlarge the prefix that every branch must consume. Prefix reuse amortizes that repeated branch cost but cannot remove the cost of constructing the shared prefix itself. The prefix-state audit further shows that shared reuse reduces repeated model computation by about threefold while preserving nearly all routing decisions (Table~\ref{tab:kv_cache_efficiency}).

\paragraph{Evidence interface and current escalation.}
\begin{table}[t]
    \centering
    \caption{Stage-wise ablation of PARSEE-VAD at 512sq.}
    \label{tab:stagewise_evidence}
    \small
    \setlength{\tabcolsep}{4.5pt}
    \begin{tabular}{lcccccc}
        \toprule
        Configuration
        & PA-routing
        & Score state
        & Spec. q/w
        & UCF AUC
        & MSAD AUC
        & MSAD AP \\
        \midrule

        Generated (no PA-readout)
        & -- & -- & 0.000
        & 80.115 & 88.211 & 78.071 \\

        Q2 only
        & -- & Q2 & 0.000
        & 84.872 & 90.275 & 79.073 \\

        + Q3 (no specialists)
        & -- & $L_t$ & 0.000
        & 84.825 & 90.273 & 79.978 \\

        + P3/P4 (non-routed)
        & $\times$ & $L_t$ & 2.000
        & -- & 90.245 & 81.155 \\

        + P3/P4 (PA-routed)
        & \checkmark & $L_t$ & 0.565
        & 85.071 & 90.518 & 81.970 \\

        \textbf{Full PARSEE-VAD}
        & \checkmark & $s_t$ & 0.565
        & \textbf{85.146} & \textbf{90.551} & \textbf{82.024} \\

        \bottomrule
    \end{tabular}

    \vspace{2pt}
    \parbox{\linewidth}{\footnotesize
    $L_t$ denotes the SEE current-evidence state; $s_t$ additionally includes temporal state propagation.\\
    ``Spec. q/w'' denotes the number of P3/P4 acquisitions per decision window.}
\end{table}
\begin{figure}[t]
    \centering
    \vspace{-5pt}
    \includegraphics[width=0.98\linewidth]{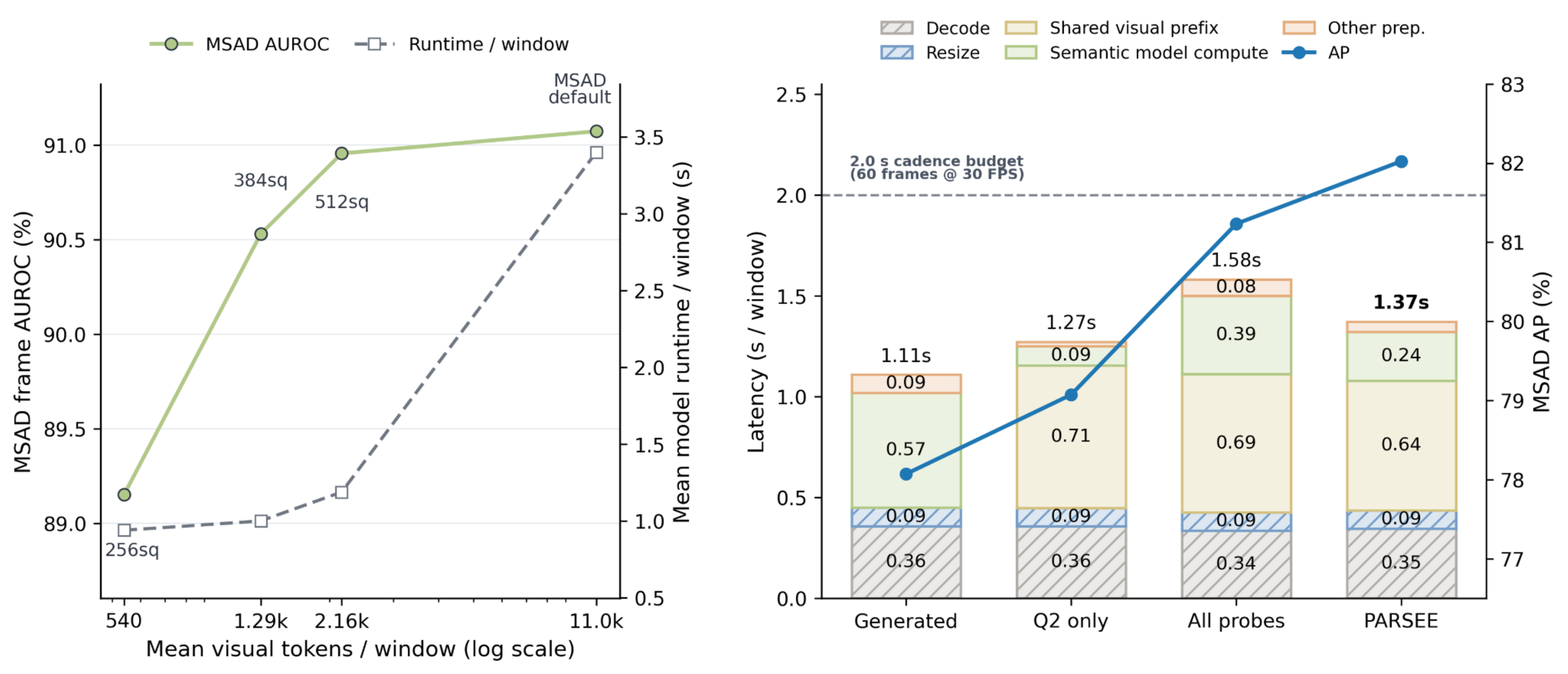}
    \caption{Efficiency analysis on MSAD. Left: visual-budget accuracy--compute trade-off. Right: latency decomposition for generated scoring, Q2-only, non-routed all-probe, and routed PARSEE inference.}
    \label{fig:efficiency}
\end{figure}
The largest early-stage gain comes from changing the evidence interface rather than adding more propositions: replacing generated 0--100 scoring with Q2 raises UCF-Crime AUC from 80.115 to 84.872 and MSAD from 88.211/78.071 to 90.275/79.073 AUC/AP. This comparison suggests that directly reading a normal/abnormal signed logit margin provides a cleaner interface than requiring the model to generate and calibrate a scalar anomaly score. Separately, the forward/reverse option-order diagnostic and the shared-prefix fidelity audit show that the resulting margin is robust to answer order and is largely preserved under prefix reuse (Tables~\ref{tab:option_order} and~\ref{tab:kv_cache_efficiency}).

Adding Q3 leaves MSAD AUC essentially unchanged while raising AP from 79.073 to 79.978. This is consistent with Q3 being a corrective dynamics proposition rather than a second coarse anomaly detector. Within SEE, a negative Q3 margin suppresses a Q2-positive interpretation that lacks visible physical development, while non-negative dynamics evidence leaves the specialist branches available to refine the event. Q3 therefore acts selectively on Q2-positive windows that lack visible dynamics, changing their relative ranking.

\paragraph{Selective specialist acquisition.}
Specialist evidence is useful, but selective acquisition is both cheaper and more effective than dense acquisition on MSAD: non-routed P3/P4 reaches 90.245/81.155 AUC/AP using two specialist queries per window, whereas PA-routing reaches 90.518/81.970 with only 0.565. The routing rule first restricts P3 to anomaly-positive, dynamically active windows, and considers P4 only for sufficiently strong Q2 cases that are not positively explained by P3; specialist evidence outside the routed subset is therefore not admitted to the same SEE current-evidence update. The runtime decomposition reflects both effects: routing lowers semantic continuation cost relative to all-probe acquisition while AP increases rather than decreases. On MSAD, PAR also outperforms random, periodic, magnitude-based, and generic-uncertainty allocation at the same 0.565-query budget (Table~\ref{tab:routing_allprobe_current}). A separate UCF-Crime diagnostic shows the same trend under its matched 0.527-query budget (Table~\ref{tab:routing_transfer}).

\paragraph{Temporal continuity.}
Temporal maintenance is sparse in practice, activating on only 1.1--2.9\% of decisions across the three datasets, compared with 23.8--43.8\% for current-evidence updates. Its aggregate increment is correspondingly small: the paired bootstrap is positive on UCF-Crime and XD-Violence, while the MSAD interval overlaps zero. The temporal stage therefore functions as a targeted continuity correction rather than a primary source of aggregate performance gain. Together with the stage-wise gains above, these statistics indicate that most score changes are determined by current-window evidence, with temporal state used for a small subset of continuity failures. The sparsity follows from the correction-carry and continuity gates defined in Sec.~3.2. Unlike SEE, the matched MA/EMA/Max baselines modify the score trajectory without event-compatibility gating (Tables~\ref{tab:stage_activity}, \ref{tab:paired_bootstrap}, and~\ref{tab:temporal_continuity_sensitivity}).

\subsection{Mechanistic Analysis}
\begin{figure}[t]
    \centering
    \includegraphics[width=\linewidth]{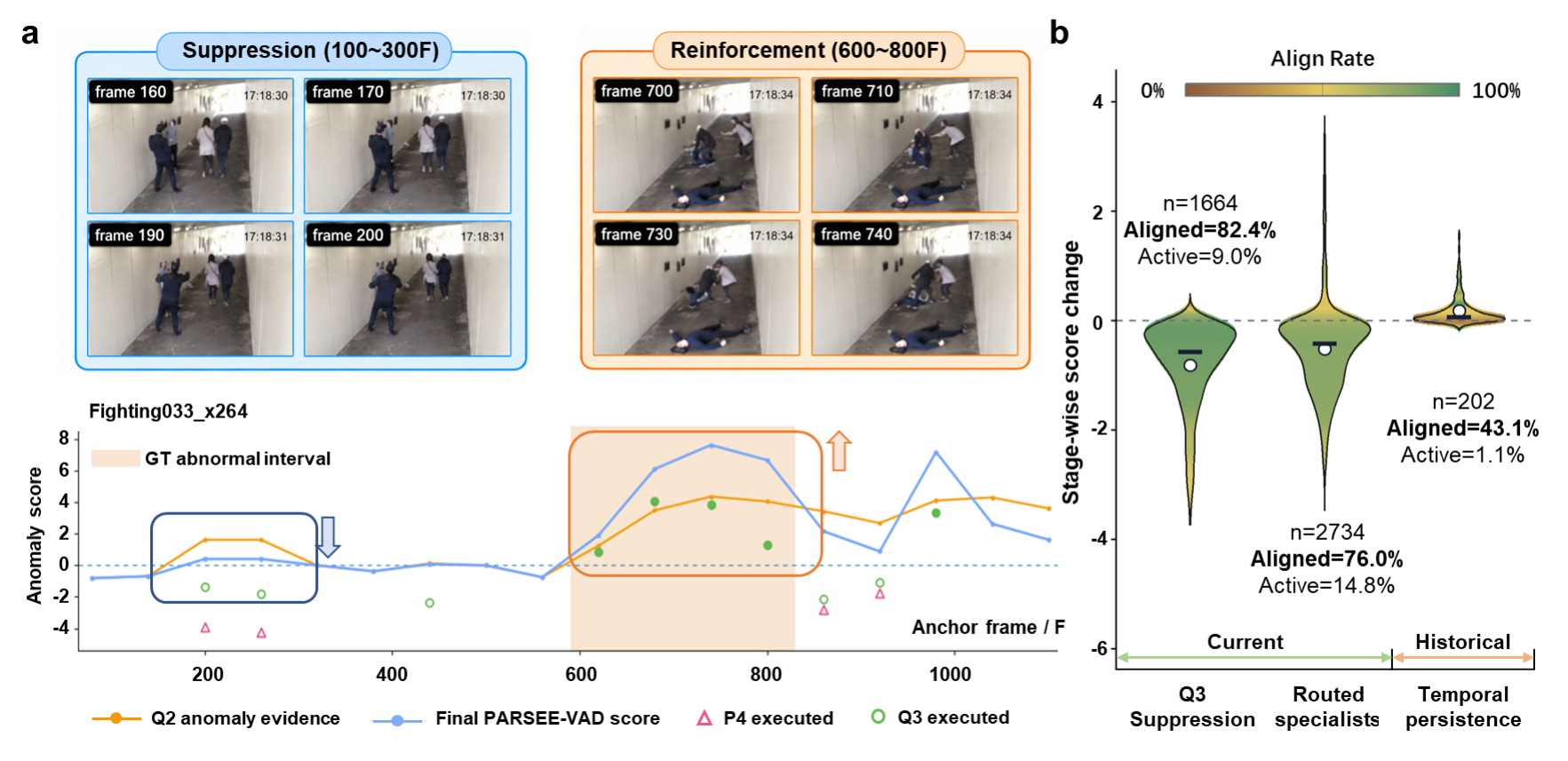}
    \caption{Mechanistic analysis. (a) Representative suppression and reinforcement cases with routed P3/P4 evidence and PARSEE-VAD scores. (b) Distributions of nonzero score-state adjustments. Active is the fraction of decisions with a nonzero adjustment, $n$ counts active adjustments, and Aligned is the fraction of active adjustments whose sign moves the score in the local ground-truth-consistent direction.}
    \label{fig:mechanism}
\end{figure}

Figure~\ref{fig:mechanism} examines where the three score-state adjustments occur. Q3 suppression and routed specialist corrections are both more frequent and more locally aligned with the benchmark labels than temporal persistence, while the temporal branch is activated only rarely. In the suppression example, Q3 reduces an early Q2-positive peak that lacks matching physical development; in the reinforcement example, routed specialist evidence strengthens the score over the labeled fighting interval. Temporal persistence appears much less often and is reserved for continuity valleys rather than routine score shaping. This coupling also reveals a failure mode: a negative Q3 margin both suppresses the current score and prevents specialist acquisition. Persistent low-dynamics hazards can therefore be penalized twice, while subtle interactions can remain unresolved when the frozen backbone fails to expose sufficient current evidence. Additional category-level and failure analyses appear in Appendices~\ref{app:additional}--\ref{app:diagnostics}.

\section{Conclusion}
PARSEE-VAD separates semantic evidence acquisition from score-state evolution for training-free online VAD. PAR reduces dependence on generated scalar scoring by exposing coarse-to-specific signed proposition evidence from a shared visual-prefix state and conditionally acquiring specialist propositions. SEE then maps that evidence into the score domain and carries only bounded score-derived state across causal decisions, avoiding repeated transmission of high-dimensional textual or visual history. Across three primary real-world benchmarks and an additional UBnormal transfer evaluation, this separation provides a compact causal interface between how present evidence is read and what evidence is allowed to persist.

\subsection*{AI use statement}
In this work, OpenAI Codex and ChatGPT were used for coding and implementation assistance, exploratory methodological discussion and feedback on experimental design, literature exploration, interpretation and analysis of experimental results, language editing, drafting support, and pre-submission reviewer-style critique. All reported quantitative results were obtained from actual program executions. The authors inspected and verified AI-assisted code, methodological suggestions, analyses, experimental results, claims, and manuscript content, and take responsibility for the final work.

\subsection*{Ethics statement}
PARSEE-VAD is developed for research on video anomaly detection and may support safety-monitoring applications. Surveillance technology carries risks including privacy infringement, false alarms, distributional bias, and inappropriate use. Our experiments use established public VAD benchmarks and focus on algorithmic evaluation. An anomaly score should not be interpreted as a determination of intent or culpability; real-world deployment requires human oversight, appropriate access control, privacy safeguards, and evaluation for the deployment domain.

\subsection*{Reproducibility statement}
The appendix provides the exact proposition prompts and inference equations together with the fixed scalar configuration, routing and temporal controls, score-availability analysis, prefix-state fidelity study, runtime protocol, backbone replays, and decision-stream integrity checks.

\bibliographystyle{iclr2027_conference}
\bibliography{reference}

\appendix
\section{Full Proposition Prompts}
\label{app:prompts}
The following prompts are copied from the evaluation configuration used by the current pipeline. For every proposition, both the original and reversed answer orders are evaluated to reduce sensitivity to option position.

\subsection{Q2: Overall anomaly evidence}
\textbf{Forward.}
\begin{verbatim}
Based only on the visual evidence available up to the current time,
classify the current event.

A. Normal
B. Abnormal

Answer with A or B only.
\end{verbatim}
\textbf{Reverse.}
\begin{verbatim}
Based only on the visual evidence available up to the current time,
classify the current event.

A. Abnormal
B. Normal

Answer with A or B only.
\end{verbatim}

\subsection{Q3: Visible physical dynamics}
\textbf{Forward.}
\begin{verbatim}
You are comparing ordered frames from the same camera.

Judge only directly visible physical change across time.
Do not judge whether the scene is normal, abnormal, suspicious,
dangerous, or important.
Do not infer motion or change merely because a single frame suggests
that an event may be occurring.

A positive answer requires temporal visual evidence across the ordered
frames showing that a person, object, interaction, or physical state
actually changes over time.

An already-existing condition does not count as active physical change
by itself. Examples include fire already burning, smoke already present,
a person already lying down, an already-damaged object, debris,
darkness, or a crowd already gathered.

Global camera motion, illumination changes, compression artifacts, or
other imaging changes do not count as the target physical dynamics.

Question: Do these ordered frames directly demonstrate clear physical
change over time?

A. No. They mainly show an already-existing state or appearance,
without clear directly demonstrated physical change over time.
B. Yes. They directly demonstrate clear physical change over time in
people, objects, interactions, or physical state.

Answer with only A or B.
\end{verbatim}
\textbf{Reverse.} The same instruction is used with the final A/B answer order reversed (Yes/No instead of No/Yes), matching the evaluation configuration.

\subsection{P3: Forceful human interaction}
\textbf{Forward.}
\begin{verbatim}
Judge only directly visible physical interaction between people across
the ordered observation.

Do not judge whether the scene is normal, abnormal, suspicious,
dangerous, or criminal.
Do not infer physical conflict merely because people are running,
gathering, pointing, waving, arguing, facing each other, or moving
around quickly.

Question: Is there clear directly visible forceful physical interaction
between people, such as hitting, pushing, grabbing, wrestling,
restraining, or another sustained physical struggle involving actual
bodily contact?

Ordinary walking or running, people gathering nearby, pointing, waving,
gesturing, arguing-like movements, or moving around without clear
forceful bodily contact do not count.

A. No. No clear forceful physical interaction between people is directly
visible.
B. Yes. Clear forceful physical interaction between people is directly
visible.

Answer with only A or B.
\end{verbatim}
\textbf{Reverse.} The same instruction is used with the final A/B answer order reversed.

\subsection{P4: Object/environment interaction}
\textbf{Forward.}
\begin{verbatim}
Judge only directly visible physical interaction with objects, property,
vehicles, or the physical environment across the ordered observation.

Do not judge whether the scene is normal, abnormal, suspicious,
dangerous, or criminal.
Do not infer an event merely from people walking, running, gathering,
pointing, waving, arguing-like gestures, or moving around.

Question: Is there a clear concrete physical interaction or event
involving an object, property, vehicle, or the environment that goes
beyond ordinary locomotion or social movement?

Positive examples include directly visible carrying or removing
property, manipulating or forcing access to an object or barrier,
climbing over or through a barrier, breaking or damaging something,
a vehicle collision or other clear vehicle-related physical event,
or another directly visible consequential physical interaction with
the environment.

Mere walking or running, crowd gathering, standing around, pointing,
waving, talking-like gestures, or ordinary vehicle movement without a
clear physical event do not count.

A. No. No clear concrete object-, vehicle-, or environment-related
physical event is directly visible.
B. Yes. A clear concrete object-, vehicle-, or environment-related
physical event is directly visible.

Answer with only A or B.
\end{verbatim}
\textbf{Reverse.} The same instruction is used with the final A/B answer order reversed.

\subsection{Generated-score comparator}
\label{app:generated_baseline}
The stagewise comparison in Table~\ref{tab:stagewise_evidence} uses a strengthened scalar-generation comparator built from the same ordered nine-frame causal observation as Q2. The model is Qwen3.5-9B with local HuggingFace generation, no system message, and the same visual preprocessing as the main 512sq setting. The user text is suffixed with \texttt{/no\_think}; the processor chat template is applied with the generation prompt enabled. The frozen run snapshots specify deterministic decoding with \texttt{do\_sample=False} and \texttt{max\_new\_tokens=8}.The display below normalizes range dashes for typesetting:
\begin{verbatim}
You are judging only what is directly visible in the ordered frames.

Assign one anomaly score from 0 to 100 for the current observation.

Interpret the score as follows:

- 0-20: clearly normal. No visible abnormal, dangerous, violent,
  destructive, or otherwise unusual event.
- 20-40: uncertain, but the visual evidence is more consistent with
  a normal explanation than an abnormal one.
- 40-60: genuinely ambiguous. Normal and abnormal explanations are
  approximately equally plausible from the visible evidence.
- 60-80: uncertain, but the visual evidence is more consistent with
  an abnormal explanation than a normal one.
- 80-100: clearly abnormal. Strong visible evidence of a dangerous,
  violent, destructive, or otherwise unusual event.

Judge only the visible evidence in these frames.
Do not infer events that are not visually supported.
Do not use outside context.
Do not explain your answer.

Answer with a single number from 0 to 100.
\end{verbatim}
The parser takes the first integer or decimal number in the response and clamps it to $[0,100]$; a response with no number is treated as a parse failure. All 3,607 MSAD windows and all 18,442 UCF-Crime windows parse successfully. For Table~\ref{tab:stagewise_evidence}, we evaluate this raw generated scalar \emph{before} the historical workflow propagation step and apply the same completed-interval frame mapping used for Q2. This gives 80.115 AUC on UCF-Crime and 88.211/78.071 AUC/AP on MSAD. The corresponding decision-anchor values are 79.778/30.583 and 87.572/78.333, while strict release-time availability gives 79.127/29.036 and 85.658/75.701, respectively. 

The raw outputs make the reported comparator metrics reproducible, but we do not claim bit-identical regeneration from the model weights: checkpoint revision/hash, tokenizer/processor hashes, library versions, and checkpoint-level generation defaults such as beam and EOS/PAD configuration were not frozen in the experiment snapshot. Parameters not explicitly passed to generation are therefore left unspecified rather than reconstructed after the fact.

\section{Exact Online Inference Operators}
\label{app:algorithm}

\subsection{PA-readout}
For every proposition query, the forward and reversed A/B answer orders are evaluated from the same reusable visual-prefix state. As in Eq.~\ref{eq:abmargin}, define $m_{k,t}^{F}=z_B^F-z_A^F$ and $m_{k,t}^{R}=z_A^R-z_B^R$; the signed proposition margin is
\begin{equation}
 \ell_k^{(t)}=\frac{m_{k,t}^{F}+m_{k,t}^{R}}{2}.
\end{equation}
Q2 and Q3 are always read to form the coarse current-window evidence, while P3 and P4 are specialist propositions whose logits are materialized only when requested by PA-routing. The exact prompt semantics are listed in Appendix~\ref{app:prompts}.

\subsection{PA-routing}
The deployed routing decisions are
\begin{align}
 r_3^{(t)} &= \mathbb{I}[\ell_{q_2}^{(t)}>0\wedge \ell_{q_3}^{(t)}\ge0],\\
 r_4^{(t)} &= \mathbb{I}[r_3^{(t)}=1\wedge \ell_{q_2}^{(t)}\ge1\wedge \ell_{p_3}^{(t)}\le0].
\end{align}
These gates decide whether P3/P4 logits are acquired. The zero thresholds follow the signed-margin semantics of Q2/Q3, whereas the more specific P4 branch uses the stricter $\ell_{q_2}^{(t)}\ge1$ requirement as a stronger entry prior. Routing uses no temporal state and ends with the set of available proposition margins.

\subsection{SEE: current evidence escalation}
Let $\vec{\ell}_t=(\ell_{q_3}^{(t)},\ell_{p_3}^{(t)},\ell_{p_4}^{(t)})$. For the compact notation used in Fig.~\ref{fig:framework}, define the complete routed proposition correction
\begin{align}
 f(\tanh(\vec{\ell}_t))
 =[\ell_{q_2}^{(t)}]_+\operatorname{clip}\Big(&-\tanh([-\ell_{q_3}^{(t)}]_+)
 +r_3^{(t)}\tanh(\ell_{q_3}^{(t)})\tanh(\ell_{p_3}^{(t)})\notag\\
 &+r_4^{(t)}\tanh(\ell_{q_3}^{(t)})\tanh(\ell_{p_4}^{(t)}),-1,1\Big),
\end{align}
and
\begin{equation}
L_t=\ell_{q_2}^{(t)}+\alpha f(\tanh(\vec{\ell}_t)),
\qquad \alpha=0.75.
\end{equation}
This memoryless operator is the first stage of SEE: PAR determines which specialist margins are available, and SEE maps the resulting signed evidence into the current-evidence state $L_t$. Because P4 is acquired only after $\ell_{p_3}^{(t)}\le0$, rejection of the P3 human-interaction hypothesis can offset positive P4 environmental evidence; Appendix~\ref{app:diagnostics} reports the positive-only alternative.

\subsection{SEE: recent correction carry}
Let $\Delta_t=L_t-\ell_{q_2}^{(t)}$. The recent-correction stage stores only the previous \emph{raw} current-evidence correction $\Delta_{t-1}$, never an inherited rescue. With $\rho_c=0.5$,
\begin{align}
 c_t &= \mathbb{I}[\Delta_t<0]\min\!\left(\rho_c[\Delta_{t-1}]_+,-\Delta_t\right),\\
 x_t &= \ell_{q_2}^{(t)}+\Delta_t+c_t=L_t+c_t.
\end{align}
When $\Delta_t<0$, this construction guarantees $x_t\le \ell_{q_2}^{(t)}$. The state then stores the current raw $\Delta_t$ for the next decision.

\subsection{SEE: historical support and temporal escalation}
Historical support uses the previous $H=2$ event states $x$ together with the coarse Q2 margins to test event continuity. Define
\begin{equation}
h_t=\frac{1}{H}\sum_{j=1}^{H}[x_{t-j}]_+,
\end{equation}
and the continuity gate
\begin{equation}
 g_t=\mathbb{I}\!\left[
 \bigwedge_{j=1}^{H}x_{t-j}>0\;\wedge\;
 \ell_{q_2}^{(t)}>0\;\wedge\;
 \ell_{q_2}^{(t)}\ge\eta \ell_{q_2}^{(t-1)}\;\wedge\;
 x_t<\tau h_t
 \right],
\end{equation}
with $\tau=0.2$ and $\eta=0.5$. The condition $\ell_{q_2}^{(t-1)}>0$ is not repeated: under the bounded current operator it is already implied by $x_{t-1}>0$. The final score is
\begin{equation}
s_t=
\begin{cases}
 \tau h_t, & g_t=1,\\[3pt]
 x_t, & g_t=0.
\end{cases}
\label{eq:see_exact}
\end{equation}
The gate itself requires $\ell_{q_2}^{(t)}>0$. The clipped routed-fusion factor lies in $[-1,1]$, so under this condition $f(\tanh(\vec{\ell}_t))=\ell_{q_2}^{(t)}\phi_t$ for some $\phi_t\in[-1,1]$. With $\alpha=0.75$, $L_t=\ell_{q_2}^{(t)}(1+\alpha\phi_t)\ge0.25\ell_{q_2}^{(t)}>0$; the correction carry is nonnegative, so $x_t>0$ as well. Thus a gated negative-valley case is algebraically unreachable and no separate negative-valley decay parameter is part of the current operator. Crucially, $x_t$ rather than $s_t$ is appended to the SEE history, so an intervention cannot recursively create the evidence that supports a later intervention.

\section{Additional Implementation Details}
\label{app:implementation}
\subsection{Shared prefix state}
The same current observation is used by Q2/Q3/P3/P4. A reusable visual-prefix state is constructed once, and proposition-specific text prefills branch from that shared state. This reuse makes independent probing computationally practical because every proposition can branch from the same visual evidence without inheriting another query's text. We refer to this as \emph{prefix-state reuse}. In the Transformers implementation used here, Qwen3.5 is a hybrid decoder: full-attention blocks maintain standard K/V states while Gated-DeltaNet blocks maintain convolutional/recurrent state. Branching at the multimodal visual/text boundary therefore reuses a heterogeneous model state, not only a standard transformer K/V tensor, and exact equivalence to an independent full-prompt forward is not guaranteed by ordinary KV-prefix arguments. We do not attribute the observed drift to one unverified positional mechanism; instead, we treat reuse as an implementation approximation and measure its proposition-logit and routing deviation directly.

\begin{table*}[t]
    \centering
\caption{Shared-prefix reuse: efficiency, routing fidelity, and score impact on MSAD.}
\label{tab:kv_cache_efficiency}
\small
\setlength{\tabcolsep}{4.2pt}
\renewcommand{\arraystretch}{1.05}

\textbf{Panel A: efficiency and proposition-logit fidelity}\par\vspace{2pt}
\begin{tabular}{lrrrr}
\toprule
Inference strategy
& \shortstack[c]{Model\\s/window $\downarrow$}
& \shortstack[c]{Total\\s/window $\downarrow$}
& \shortstack[c]{Model\\speedup $\uparrow$}
& \shortstack[c]{Mean headwise\\logit MSE $\downarrow$} \\
\midrule
Independent full forward
& 2.494 & 4.006 & 1.00$\times$ & 0.0000 \\
\textbf{Native prefix reuse (ours)}
& \textbf{0.839} & \textbf{2.383} & \textbf{2.97$\times$} & 0.0214 \\
\bottomrule
\end{tabular}

\vspace{6pt}
\textbf{Panel B: routing preservation}\par\vspace{2pt}
\begin{tabular}{lrr}
\toprule
Decision & Agreement (\%) & Flips \\
\midrule
$q_2>0$ & 99.31 & 25 / 3,607 \\
$q_3\ge0$ & 99.00 & 36 / 3,607 \\
$q_2\ge1$ & 99.09 & 33 / 3,607 \\
$p_3\le0$ (both available) & 99.75 & 3 / 1,185 \\
P3 route & 99.03 & 35 / 3,607 \\
P4 route & 98.92 & 39 / 3,607 \\
Complete route path & \textbf{98.28} & 62 / 3,607 \\
\bottomrule
\end{tabular}

\vspace{6pt}
\textbf{Panel C: score replay on the controlled cache trace}\par\vspace{2pt}
\begin{tabular}{lrr}
\toprule
Inference strategy & AUC (\%) $\uparrow$ & AP (\%) $\uparrow$ \\
\midrule
Independent full forward & 90.611 & 82.063 \\
Native prefix reuse & 90.539 & 81.960 \\
$\Delta$ (native $-$ full) & $-0.072$ & $-0.103$ \\
\bottomrule
\end{tabular}

\vspace{6pt}
\textbf{Panel D: temporal contribution by execution mode}\par\vspace{2pt}
\begin{tabular}{lrrrr}
\toprule
Inference strategy & \multicolumn{2}{c}{Current-evidence state $L_t$} & \multicolumn{2}{c}{Final score $s_t$} \\
\cmidrule(lr){2-3} \cmidrule(lr){4-5}
& AUC (\%) & AP (\%) & AUC (\%) & AP (\%) \\
\midrule
Independent full forward & 90.591 & 82.028 & 90.611 & 82.063 \\
Native prefix reuse & 90.515 & 81.924 & 90.539 & 81.960 \\
\bottomrule
\end{tabular}

\vspace{2pt}
\parbox{0.97\textwidth}{\footnotesize Panel A uses the archived matched traces in \texttt{cache/}. Mean headwise MSE is the unweighted mean of Q2/Q3/P3/P4 MSEs over rows where each head is available; ``both available'' denotes anchors where P3 was acquired in both traces. The specialist workload differs by eight calls over 3,607 windows (0.56584 versus 0.56363 probes/window). Panel C is a controlled cache-fidelity replay rather than the canonical headline stream. Panel D compares $L_t$ and $s_t$ within the same matched traces.}

\end{table*}

Table~\ref{tab:kv_cache_efficiency} characterizes the approximation introduced by branching from a shared hybrid prefix state. Native reuse reduces synchronized model time by about threefold while keeping proposition-logit drift small enough that almost all routing paths are preserved. Replaying the same SEE score-state operator on the matched traces shows that this approximation mainly perturbs absolute score geometry rather than the direction of the temporal update. Final optimized deployment latency is reported separately in Appendix~\ref{app:runtime}.

\subsection{Hardware and runtime measurement}
\label{app:runtime}
All final runtime variants were measured on the same server GPU model, NVIDIA RTX 5880 Ada Generation (49,140 MiB VRAM per GPU), with Intel Xeon Gold 6526Y CPUs and Ubuntu 22.04.4. The runtime environment uses Python 3.11.15, PyTorch 2.9.1+cu128, Transformers 5.13.1, CUDA 12.8, and bfloat16 Qwen3.5-9B inference. The installed stack uses the Transformers/Qwen default attention backend: \texttt{flash\_attn} is unavailable, and no explicit \texttt{attn\_implementation} is configured.

Each measured variant processes one decision window at a time and one complete MSAD pass of 3,607 windows from 360 videos. Runtime uses \texttt{time.perf\_counter()} with explicit CUDA synchronization around visual-prefix and proposition-tail model sections. End-to-end time includes frame decoding, resizing, shared visual prefill, proposition continuations, processor/tokenization work occurring inside the measured path, and Python/orchestration residual. No warm-up windows were excluded from the final full-set summaries. Q2-only, dense all-probe, and routed PARSEE-VAD were executed as separate one-GPU processes on GPU indices 4, 5, and 7, respectively; all cards are the same RTX 5880 Ada model. CPU-core/NUMA/storage affinity was not explicitly isolated, so the synchronized model-section timing is the cleaner algorithmic compute comparison; end-to-end timing may also contain host-side contention and is reported as a system measurement rather than a hardware-invariant quantity. The controlled prefix-fidelity harness in Table~\ref{tab:kv_cache_efficiency} uses separate instrumentation, so its total-time measurements are interpreted within that harness rather than mixed with the deployment timing above.

\begin{table}[t]
\centering
\caption{Full-set MSAD runtime at 512sq.}
\label{tab:runtime_fullset}
\small
\setlength{\tabcolsep}{4.2pt}
\begin{tabular}{lrrrrr}
\toprule
Variant & Specialists / win. & Model mean & Total mean & Total med. & Total P95 \\
 &  & (s) & (s) & (s) & (s) \\
\midrule
Q2 only & 0.000 & 0.802 & 1.271 & 1.190 & 1.687 \\
Dense all-probe & 2.000 & 1.073 & 1.581 & 1.499 & 1.988 \\
PARSEE routed & 0.565 & 0.884 & 1.372 & 1.378 & 1.741 \\
\bottomrule
\end{tabular}
\vspace{2pt}
\parbox{0.96\linewidth}{\footnotesize Each variant is one complete 3,607-window pass on an RTX 5880 Ada GPU of the same server. ``Model'' is the synchronized visual-prefix plus proposition-tail section; ``Total'' is end-to-end wall clock per window. Relative to dense all-probe acquisition, routed PARSEE uses 71.7\% fewer specialist queries, 17.6\% lower mean model time, and 13.2\% lower mean total time.}
\end{table}

The dense-to-routed difference is 13.2\% in end-to-end mean and 17.6\% in synchronized model-section mean. These values are intentionally reported separately from the 71.7\% reduction in specialist-query count. The older 1.91/2.09 s/window values used in intermediate drafts were provisional estimates and are superseded by these direct full-set measurements. The runtime configuration, measurement protocol, and full-set timing results are reported above and in Table~\ref{tab:runtime_fullset}.

\subsection{Low-latency operating point}
\label{app:low_latency}
We additionally test a lower-latency deployment setting that changes only the observation budget and decision cadence. The canonical nine-frame observation $[-80,-70,\ldots,0]$ with a 60-frame stride is replaced by four frames $[-30,-20,-10,0]$ with a 30-frame stride, while Qwen3.5-9B, 512sq preprocessing, PA-routing, SEE, and all inference-rule constants remain unchanged.

\begin{table*}[t]
    \centering
    \small
    \setlength{\tabcolsep}{5pt}
    \begin{tabular}{llcccc}
        \toprule
        Dataset & Operating point & AUC (\%) $\uparrow$ & AP (\%) $\uparrow$ & Mean E2E (s) $\downarrow$ & P95 E2E (s) $\downarrow$ \\
        \midrule
        MSAD & 9F / 60F stride & 90.551 & 82.024 & 1.372 & 1.741 \\
        MSAD & 4F / 30F stride & 90.423 & 83.059 & 0.753 & 0.931 \\
        UCF-Crime & 9F / 60F stride & 85.146 & -- & -- & -- \\
        UCF-Crime & 4F / 30F stride & 84.567 & -- & 0.485 & 0.658 \\
        \bottomrule
    \end{tabular}
    \caption{Canonical and low-latency PARSEE-VAD operating points. The 4F/30F rows use the same model, 512sq budget, PA-routing, and SEE rules as the main system. Matching full-set UCF-Crime timing for the canonical 9F/60F setting was not retained, so those timing cells are left blank.}
    \label{tab:low_latency}
\end{table*}

The 4F/30F runs cover 7,354 MSAD decisions from 360 videos and 36,917 UCF-Crime decisions from 290 videos. On MSAD, the shorter observation reduces mean/P95 end-to-end time from 1.372/1.741 s to 0.753/0.931 s while changing AUC/AP from 90.551/82.024 to 90.423/83.059. On UCF-Crime, it reaches 84.567 AUC with 0.485/0.658 s mean/P95 processing, compared with 85.146 AUC for the canonical setting. Because temporal span and cadence change together, we treat this result as a deployment operating point rather than an isolated ablation of context length or observation overlap.

\subsection{Frame sampling, score availability, and integrity}
The reported 512sq runs contain all expected decisions: 18,442 for UCF-Crime, 39,418 for XD-Violence, and 3,607 for MSAD. Nominal observation offsets are $[-80,-70,\ldots,0]$ relative to each anchor; indices before stream onset are clamped to frame 0, and the dataset/run decision anchors are preserved (including a small number of MSAD videos with a shorter regular stride). SEE starts with empty score and Q2 histories and therefore cannot activate its $H=2$ historical gate until two preceding $x$-states are available. There are no duplicate video--anchor pairs, non-empty failure statuses, or decision-index gaps. Independent reconstruction of the official labels gives 0/18,442 UCF-Crime anchor mismatches, 0/39,418 XD-Violence mismatches, and 0/3,607 MSAD mismatches. XD-Violence frame counts cover 800 videos and 2,335,801 continuous frame indices. Complete videos are assigned to individual workers so temporal state never crosses shards.

\subsection{Fixed inference parameters}
\label{app:fixedparams}
All reported evaluations use one scalar configuration: SEE current-escalation strength $\alpha=0.75$; one-step correction-carry coefficient $\rho_c=0.5$; SEE horizon $H=2$; valley ratio $\tau=0.2$; and Q2 continuity ratio $\eta=0.5$. These choices encode simple operator priors: signed zero gates follow the Q2/Q3 margin semantics, the P4 $\ell_{q_2}^{(t)}\ge1$ gate requires stronger coarse support for the narrowest specialist, $\alpha=0.75$ limits the current proposition correction to a substantial but bounded fraction of positive Q2 evidence, and the temporal values encode short-horizon continuity over the preceding two decisions. The same values are used for all reported datasets; sensitivity to nearby settings is reported below.

\subsection{Current escalation and routing sensitivity}
\begin{table*}[t]
\centering
\caption{Sensitivity of PAR acquisition and SEE current escalation.}
\label{tab:scalar_sensitivity}
\small
\setlength{\tabcolsep}{5pt}
\begin{tabular}{lrrrr}
\toprule
\multicolumn{5}{l}{\textbf{(a) Current-escalation strength $\alpha$}}\\
$\alpha$ & UCF AUC & MSAD AUC & MSAD AP & Comment \\
\midrule
0.25 & 85.052 & 90.557 & 81.266 & weaker local update \\
0.50 & 85.132 & \textbf{90.632} & 82.085 &  \\
0.625 & 85.144 & 90.602 & \textbf{82.101} &  \\
\textbf{0.75 (used)} & \textbf{85.146} & 90.551 & 82.024 & deployed \\
0.875 & 85.134 & 90.457 & 81.851 &  \\
1.00 & 84.865 & 89.640 & 80.628 & stronger local update \\
\midrule
\multicolumn{5}{l}{\textbf{(b) P4 entry gate on MSAD}}\\
Gate & P4 calls & Specialists / win. & MSAD AUC & MSAD AP \\
\midrule
0.0 & 1018 & 0.615 & 90.508 & 81.962 \\
0.5 & 935 & 0.592 & 90.518 & 81.976 \\
\textbf{1.0 (used)} & \textbf{837} & \textbf{0.565} & \textbf{90.551} & \textbf{82.024} \\
1.5 & 723 & 0.534 & 90.611 & 82.131 \\
2.0 & 622 & 0.506 & 90.630 & 82.188 \\
\bottomrule
\end{tabular}
\vspace{2pt}
\parbox{0.97\textwidth}{\footnotesize The response is broad rather than sharply peaked around the deployed setting: nearby $\alpha$ values slightly improve individual MSAD metrics, while stricter P4 gates improve MSAD but show the opposite trend on the archived UCF-Crime dense-probe subset.}
\end{table*}

Nearby SEE current-escalation strengths produce a broad performance plateau: $\alpha=0.50$ gives slightly higher MSAD AUC, while $\alpha=0.625$ gives slightly higher AP than the deployed $\alpha=0.75$. P4 entry-threshold sweeps show a similarly broad operating region; stricter finite thresholds improve MSAD while the archived UCF-Crime dense-probe subset favors a looser gate. The available sweep therefore characterizes threshold robustness over the measured finite gates, while the absence of a full-set P3-only boundary leaves P4 necessity as a separate question. Across the measured settings, performance varies gradually rather than peaking sharply at the deployed values.

\section{Additional Analyses}
\label{app:additional}
\subsection{Backbone robustness}
\begin{table*}[htbp]
\centering
\caption{Backbone robustness on MSAD at 512sq.}
\label{tab:backbone_robustness}
\small
\setlength{\tabcolsep}{6.5pt}
\begin{tabular}{lccccccc}
\toprule
&& \multicolumn{2}{c}{Q2 only} & \multicolumn{2}{c}{PARSEE-VAD} & \multicolumn{2}{c}{$\Delta$} \\
\cmidrule(lr){3-4}\cmidrule(lr){5-6}\cmidrule(lr){7-8}
Backbone & Size & AUC $\uparrow$ & AP $\uparrow$ & AUC $\uparrow$ & AP $\uparrow$ & AUC & AP \\
\midrule
Qwen3.5-2B & 2B & 88.499 & 77.869 & 87.514 & 78.881 & -0.985 & +1.012 \\
VideoLLaMA3-7B & 7B & 88.986 & 79.071 & 89.311 & 81.162 & +0.325 & +2.090 \\
Qwen3.5-9B & 9B & 90.275 & 79.073 & 90.551 & 82.024 & +0.276 & +2.951 \\
\bottomrule
\end{tabular}
\vspace{2pt}
\parbox{0.96\textwidth}{\footnotesize
All backbones use the same nine causal frames, proposition prompts, routing thresholds, and score-state coefficients without backbone-specific retuning. $\Delta$ is the change from the corresponding Q2-only baseline. Values are frame-level MSAD metrics over the same 3,607 decision windows.}
\end{table*}

The same 3,607 MSAD decision windows are replayed under the current operator for all three backbones. The stage paths differ substantially: Qwen3.5-2B experiences aggressive Q3 suppression, VideoLLaMA3-7B obtains most of its AP gain from Q3, and Qwen3.5-9B benefits most strongly from specialist-driven current escalation. The final AP improvement across all three backbones therefore does not imply that every proposition contributes with the same sign or magnitude. In particular, the Qwen3.5-2B AUC drop shows that fixed signed-margin thresholds are not automatically calibrated across model capacities; backbone-specific margin calibration or quantile/temperature normalization is a natural extension. This cross-backbone diagnostic is limited to MSAD.

\subsection{Intervention density}
\begin{table}[t]
\centering
\caption{SEE intervention density across datasets.}
\label{tab:stage_activity}
\small
\setlength{\tabcolsep}{5.0pt}
\begin{tabular}{lrrr}
\toprule
Dataset & \shortstack{Current update\\(\%)} & \shortstack{Temporal maintenance\\(\%)} & \shortstack{Final valley\\gate (\%)} \\
\midrule
UCF-Crime   & 23.8 & 1.1 & 0.32 \\
MSAD        & 43.8 & 2.9 & 0.69 \\
XD-Violence & 40.7 & 2.9 & 0.56 \\
\bottomrule
\end{tabular}
\vspace{2pt}
\parbox{0.96\linewidth}{\footnotesize Current update counts decisions where Q3/P3/P4 evidence changes the coarse Q2 state. Only temporal maintenance and the final valley gate consult historical state; all observed valley-gate activations are positive-score recoveries.}
\end{table}

SEE has a broad memoryless current-evidence stage and a much sparser temporal stage. Current Q3/P3/P4 corrections are comparatively frequent, whereas historical maintenance activates only when a recent correction or continuity valley satisfies its gate; the final valley gate is the rarest temporal intervention.

\section{Additional Evaluation Analyses}
\label{app:diagnostics}

\paragraph{Dense and budget-matched routing controls.}
\begin{table*}[t]
\centering
\caption{MSAD routing audit on a shared 512sq proposition bank.}
\label{tab:routing_allprobe_current}
\small
\setlength{\tabcolsep}{4.5pt}
\begin{tabular}{lrrrr}
\toprule
Configuration & Specialists / win. & Total s / win. & AUC & AP \\
\midrule
\multicolumn{5}{l}{\textbf{(a) Compute gating versus evidence selection; final score $s_t$}}\\
PA-routed acquisition & 0.565 & 1.372 & \textbf{90.551} & \textbf{82.024} \\
Always compute + routed mask & 2.000 & 1.581 & \textbf{90.551} & \textbf{82.024} \\
Always compute + dense evidence & 2.000 & 1.581 & 90.280 & 81.233 \\
\midrule
\multicolumn{5}{l}{\textbf{(b) Budget-matched acquisition; current-evidence state $L_t$}}\\
Uniform random, 1,000 seeds & 0.565 & -- & $90.147\pm0.116$ & $80.312\pm0.508$ \\
Random among Q2-positive & 0.565 & -- & $90.080\pm0.121$ & $80.535\pm0.489$ \\
Fixed periodic allocation & 0.565 & -- & 90.345 & 80.683 \\
Top-$q_2$ magnitude only & 0.565 & -- & 90.144 & 81.194 \\
Generic uncertainty allocation & 0.565 & -- & 89.938 & 79.437 \\
\textbf{PAR semantic routing} & \textbf{0.565} & -- & \textbf{90.518} & \textbf{81.970} \\
\bottomrule
\end{tabular}
\vspace{2pt}
\parbox{0.97\textwidth}{\footnotesize The all-probe execution path is directly timed at 1.581 s/window over the full MSAD set. ``Always compute + routed mask'' applies the deployed PA-routing mask post hoc to the materialized dense logits, so its score is identical to routed compute by construction while its measured execution cost is that of all-probe acquisition. The random/simple controls are post-hoc counterfactuals and do not require additional model inference.}
\end{table*}

Panel A separates the cost of computing specialist evidence from the effect of admitting it into the score state. Panel B then fixes the specialist budget and changes only allocation. This makes the routing question semantic rather than purely computational: PAR spends the same budget according to the Q2/Q3/P3 proposition structure, whereas random, periodic, magnitude, or generic uncertainty policies ignore that decomposition. Table~\ref{tab:routing_transfer} tests the same allocation principle on the archived UCF-Crime dense-probe bank.

\begin{table}[t]
\centering
\caption{Cross-dataset routing diagnostic on UCF-Crime.}
\label{tab:routing_transfer}
\small
\setlength{\tabcolsep}{4.5pt}
\begin{tabular}{lrrr}
\toprule
Policy & Specialists / win. & AUC & AP \\
\midrule
Dense all specialists & 2.000 & 67.537 & 25.476 \\
Uniform random & 0.527 & $68.213\pm0.224$ & $28.952\pm0.654$ \\
Generic uncertainty & 0.527 & 68.090 & 31.488 \\
\textbf{PAR semantic routing} & \textbf{0.527} & \textbf{68.779} & \textbf{34.418} \\
\bottomrule
\end{tabular}
\vspace{2pt}
\parbox{0.96\linewidth}{\footnotesize The archived dense-probe subset contains 70 anomaly videos and 4,441 windows. Random allocation reports 1,000 seeds; all non-dense policies use the same specialist budget. PAR exceeds 995/1,000 random runs in AUC and all 1,000 in AP.}
\end{table}

\paragraph{Decision-stream integrity.}
The three primary 512sq evaluations contain all expected causal decisions: 18,442 for UCF-Crime, 39,418 for XD-Violence, and 3,607 for MSAD, with no duplicate video--anchor pairs, non-empty failure statuses, or decision-index gaps. Independent reconstruction of the official labels gives zero anchor-label mismatches on these three datasets. The additional UBnormal transfer run contains 1,649 decisions over all 211 official test videos; its sanitized frame-level GT contains 92,640 frames (49,850 anomalous and 42,790 normal). Complete videos are assigned to individual workers so temporal state never crosses video or shard boundaries, and the benchmark adapter expands each completed interval exactly once.

\subsection{UBnormal comparison provenance}
\label{app:ubnormal_provenance}
Table~\ref{tab:main_comparison} adds UBnormal as an additional transfer benchmark. To avoid conflating native reports with secondary comparison values, we record the source of every non-empty UBnormal entry. Sultani et al. (50.30), RTFM (64.94), STPrompt (63.98), OVVAD (62.94), Holmes-VAU (56.77), LAVAD (51.06), VADTree (65.80), and PANDA (75.78) are transcribed from the UBnormal column of SphereVAD's comparison table~\citep{SphereVAD}; these are therefore treated as published comparison values rather than our reruns. In particular, the VADTree value is a secondary/reproduced result reported there rather than a native UBnormal value from the VADTree main table. SphereVAD (full) 76.46 is from the same comparison table, while SphereVAD (online) 71.62 is from its reported online deployment mode, which removes the offline test-set-dependent components~\citep{SphereVAD}. URF-HVAA 68.98 is taken from the original fixed-constant-$m$ setting~\citep{URFHVAA}; that work also reports 69.02 for its adaptive setting, but we use 68.98 because the UCF-Crime/XD-Violence/MSAD values in our URF-HVAA row are from the fixed-constant setting.

PARSEE-VAD's UBnormal value is obtained from a dedicated 512sq run on the official 211-video test split using the same frozen Qwen3.5-9B, nine-frame causal observation, 60-frame decision cadence, PA-routing, and SEE operator used in the main experiments. The frame-level ground truth reconstructed from the official test annotations contains 92,640 frames, with 49,850 anomalous and 42,790 normal frames. Applying the completed-interval mapping $(t_{k-1},t_k]$ yields a frame-level micro AUROC of 0.701769962 (70.18\%); the corresponding AP is 0.759735912. Entries shown as ``--'' are left unfilled rather than mixing in UBnormal values whose reporting source or evaluation protocol was not verified for this comparison.

\paragraph{Score-availability sensitivity.}
The main benchmark adapter assigns $s_k$ retrospectively to its completed interval $I_k$. We therefore also use an anchor-aligned causal view in which $s_k$ first enters the frame sequence at the decision anchor $t_k$ and is carried forward until the next anchor. This isolates the timing effect of interval closure from the benchmark mapping.
\begin{table*}[t]
\centering
\caption{Completed-interval and anchor-aligned score availability.}
\label{tab:online_availability_all}
\small
\setlength{\tabcolsep}{3.5pt}
\begin{tabular}{lccccc}
\toprule
& \multicolumn{1}{c}{UCF AUC} & \multicolumn{2}{c}{XD-Violence} & \multicolumn{2}{c}{MSAD} \\
\cmidrule(lr){3-4}\cmidrule(lr){5-6}
Stage & C / A & AUC C / A & AP C / A & AUC C / A & AP C / A \\
\midrule
Q2 & 84.872 / 83.953 & 92.640 / 90.774 & 76.047 / 71.465 & 90.275 / 88.118 & 79.073 / 76.257 \\
$L_t$ (current) & 85.071 / 84.148 & 92.940 / 91.041 & 78.876 / 74.015 & 90.518 / 88.242 & 81.970 / 78.697 \\
$s_t$ (final) & 85.146 / 84.208 & 92.999 / 91.082 & 79.016 / 74.104 & 90.551 / 88.236 & 82.024 / 78.669 \\
\bottomrule
\end{tabular}
\vspace{2pt}
\parbox{0.97\textwidth}{\footnotesize C assigns each score retrospectively over its completed interval; A exposes it only from the causal decision anchor onward. A does not include model processing time after the anchor.}
\end{table*}

The anchor-aligned view still abstracts away the processing time after $t_k$. For a serial implementation with measured per-window processing time $C_k$, the wall-clock release obeys $r_k=\max(t_k,r_{k-1})+C_k$, so any compute overrun is inherited by the next decision. The runtime audit provides the measured processing-time component $C_k$: the canonical MSAD operating point uses a 2-s decision interval with 1.372-s mean and 1.741-s P95 processing, while the 4F/30F operating point in Appendix~\ref{app:low_latency} uses a 1-s interval with 0.753-s mean and 0.931-s P95 processing. Together, the anchor-aligned metrics and runtime trace distinguish causal score mapping from system compute delay.

\paragraph{Video-level uncertainty of component gains.}
\begin{table*}[t]
\centering
\caption{Paired video-level bootstrap of score-state increments.}
\label{tab:paired_bootstrap}
\small
\setlength{\tabcolsep}{4pt}
\begin{tabular}{llcc}
\toprule
Dataset & Contrast & $\Delta$AUC (95\% CI) & $\Delta$AP (95\% CI) \\
\midrule
UCF-Crime & Q2 $\rightarrow L_t$ & +0.199 [$-0.098$, +0.536] & -- \\
          & $L_t \rightarrow s_t$ & +0.075 [+0.031, +0.129] & -- \\
XD-Violence & Q2 $\rightarrow L_t$ & +0.300 [$-0.099$, +0.738] & +2.828 [+0.734, +5.155] \\
            & $L_t \rightarrow s_t$ & +0.060 [+0.034, +0.088] & +0.140 [+0.074, +0.210] \\
MSAD & Q2 $\rightarrow L_t$ & +0.243 [$-0.505$, +1.446] & +2.897 [$-1.204$, +7.466] \\
     & $L_t \rightarrow s_t$ & +0.032 [$-0.008$, +0.079] & +0.054 [$-0.046$, +0.168] \\
     & dense $L_t \rightarrow$ routed $L_t$ & +0.273 [$-0.271$, +0.910] & +0.815 [$-1.224$, +3.371] \\
\bottomrule
\end{tabular}
\vspace{2pt}
\parbox{0.97\textwidth}{\footnotesize Complete videos are resampled with replacement for 5,000 paired replicates. Values are percentage-point changes with percentile 95\% intervals under completed-interval frame-level evaluation.}
\end{table*}

Paired resampling separates the current-evidence and temporal parts of the score-state path. The Q2$\rightarrow L_t$ change is largest in AP, while the additional temporal $L_t\rightarrow s_t$ increment is smaller and overlaps zero on MSAD. The wide video-level intervals also show that several small AUC changes are dominated by inter-video variation in these untrimmed benchmarks.

\paragraph{Decision-anchor stage decomposition.}
\begin{table}[!htbp]
\centering
\caption{Decision-anchor stage decomposition.}
\label{tab:decision_stage}
\small
\setlength{\tabcolsep}{4.6pt}
\begin{tabular}{lrrrrrr}
\toprule
& \multicolumn{2}{c}{Q2 only} & \multicolumn{2}{c}{PARSEE-VAD} & \multicolumn{2}{c}{$\Delta$} \\
\cmidrule(lr){2-3}\cmidrule(lr){4-5}\cmidrule(lr){6-7}
Dataset & AUC & AP & AUC & AP & AUC & AP \\
\midrule
UCF-Crime   & 84.43 & 34.52 & 84.69 & 38.67 & +0.25 & +4.14 \\
XD-Violence & 92.24 & 74.74 & 92.73 & 78.35 & +0.49 & +3.61 \\
MSAD        & 89.63 & 78.92 & 90.11 & 82.47 & +0.48 & +3.54 \\
\bottomrule
\end{tabular}
\vspace{1mm}
\footnotesize All deltas are computed from unrounded metrics.
\end{table}

The Q2-to-full changes are already visible at the causal decision anchors, so the main score-state improvement does not depend on completed-interval frame expansion. The AP change is dominated by current proposition escalation, while the additional temporal update is intentionally sparse.

\paragraph{Option-order balance.}
\begin{table*}[!htbp]
\centering
\caption{Option-order sensitivity of the Q2 proposition readout.}
\label{tab:option_order}
\small
\setlength{\tabcolsep}{5.0pt}
\begin{tabular}{lrrrrrrr}
\toprule
& \multicolumn{2}{c}{Forward only} & \multicolumn{2}{c}{Reverse only} & \multicolumn{2}{c}{Balanced} & Flip (\%) \\
\cmidrule(lr){2-3}\cmidrule(lr){4-5}\cmidrule(lr){6-7}
Dataset & AUC & AP & AUC & AP & AUC & AP & \\
\midrule
UCF-Crime   & 84.33 & 33.28 & 84.06 & 34.98 & 84.43 & 34.52 & 10.5 \\
XD-Violence & 92.06 & 73.98 & 91.85 & 73.93 & 92.24 & 74.74 & 24.1 \\
MSAD        & 89.92 & 78.79 & 89.01 & 78.83 & 89.63 & 78.92 & 7.9 \\
\bottomrule
\end{tabular}
\vspace{2pt}
\parbox{0.97\textwidth}{\footnotesize ``Balanced'' averages sign-aligned forward and reversed A/B margins. Flip is the fraction of anchors whose $>0$ decision differs between the two prompt orders.}
\end{table*}

Forward and reversed A/B margins remain strongly correlated, but their signs differ on a non-negligible subset of decisions. Averaging the sign-aligned orders is therefore used as the fixed PAR proposition readout rather than relying on an arbitrary option position.

\paragraph{Specialist evidence semantics.}
\begin{table*}[!htbp]
\centering
\caption{Specialist evidence semantics at decision anchors.}
\label{tab:fusion_semantics}
\small
\setlength{\tabcolsep}{7pt}
\begin{tabular}{l|cc|cc|cc}
\toprule
\multirow{2}{*}{Current-evidence update rule} & \multicolumn{2}{c|}{UCF-Crime} & \multicolumn{2}{c|}{XD-Violence} & \multicolumn{2}{c}{MSAD} \\
\cmidrule(lr){2-3}\cmidrule(lr){4-5}\cmidrule(lr){6-7}
& AUC & AP & AUC & AP & AUC & AP \\
\midrule
Signed specialist correction (main) & 84.69 & 38.67 & 92.73 & 78.35 & 90.11 & 82.47 \\
Positive-only specialist reinforcement & 84.54 & 36.09 & 93.01 & 78.46 & 90.10 & 82.49 \\
\bottomrule
\end{tabular}
\vspace{2pt}
\parbox{0.96\textwidth}{\footnotesize Positive-only specialist scoring preserves negative P3 for P4 routing but clips negative P3/P4 score contributions to zero. Values are decision-anchor diagnostics and should not be compared numerically with completed-segment benchmark metrics.}
\end{table*}

The signed evidence rule lets a routed specialist provide either supporting or rejecting evidence about its semantic hypothesis. The counterfactual exposes a genuine trade-off rather than a universally beneficial sign choice: signed suppression is better on UCF-Crime at decision anchors, whereas positive-only scoring is slightly better on XD-Violence and MSAD. In particular, because P4 is routed only after a non-positive P3 result, the deployed signed rule can offset positive environmental evidence with rejection of the human-interaction hypothesis. We retain the deployed signed rule and report the positive-only alternative as a semantic diagnostic.

\paragraph{Matched causal temporal baselines.}
Table~\ref{tab:temporal_continuity_sensitivity}, Panel A, starts from the same current-evidence state $L_t$ and changes only the temporal rule. Generic smoothing can exploit persistence on some datasets but can also reorder scores broadly; the temporal part of SEE instead acts only when its short-horizon state conditions are satisfied.

\paragraph{Temporal-continuity sensitivity.}
\begin{table*}[!htbp]
\centering
\caption{Temporal maintenance and parameter sensitivity.}
\label{tab:temporal_continuity_sensitivity}
\small
\setlength{\tabcolsep}{5.0pt}
\renewcommand{\arraystretch}{1.04}

\textbf{Panel A: matched causal temporal rules}\par\vspace{2pt}
\begin{tabular}{lccccc}
\toprule
Temporal rule & UCF AUC & XD AUC & XD AP & MSAD AUC & MSAD AP \\
\midrule
Current-evidence state $L_t$ & 85.071 & 92.940 & 78.876 & 90.518 & 81.970 \\
Causal MA3 & 85.630 & 92.651 & 77.606 & 89.405 & 79.504 \\
Causal EMA ($\beta=0.5$) & 85.987 & \textbf{93.348} & \textbf{79.620} & 89.752 & 80.290 \\
Causal Max-3 & 85.560 & 92.221 & 75.230 & 89.891 & 79.325 \\
\textbf{Final score $s_t$} & 85.146 & 92.999 & 79.016 & \textbf{90.551} & \textbf{82.024} \\
\bottomrule
\end{tabular}

\vspace{6pt}
\textbf{Panel B: temporal-parameter sensitivity at decision anchors}\par\vspace{2pt}
\begin{tabular}{lcccccc}
\toprule
& \multicolumn{2}{c}{UCF-Crime} & \multicolumn{2}{c}{XD-Violence} & \multicolumn{2}{c}{MSAD} \\
\cmidrule(lr){2-3}\cmidrule(lr){4-5}\cmidrule(lr){6-7}
Setting & AUC & AP & AUC & AP & AUC & AP \\
\midrule
base ($\rho_c=0.5,H=2,\tau=0.2,\eta=0.5$) & 84.69 & 38.67 & 92.73 & 78.35 & 90.11 & 82.47 \\
$\tau=0.10$ & 84.67 & 38.61 & 92.72 & 78.34 & 90.10 & 82.46 \\
$\tau=0.30$ & 84.71 & 38.77 & 92.74 & 78.38 & 90.09 & 82.44 \\
$\eta=0.25$ & 84.73 & 38.75 & 92.73 & 78.35 & 90.08 & 82.43 \\
$\eta=0.75$ & 84.67 & 38.63 & 92.73 & 78.35 & 90.10 & 82.46 \\
$H=1$ & 84.68 & 38.63 & 92.72 & 78.34 & 90.11 & 82.47 \\
$H=3$ & 84.69 & 38.68 & 92.73 & 78.36 & 90.09 & 82.44 \\
$\rho_c=0.25$ & 84.67 & 38.58 & 92.73 & 78.34 & 90.11 & 82.48 \\
$\rho_c=0.75$ & 84.69 & 38.71 & 92.73 & 78.36 & 90.10 & 82.44 \\
\bottomrule
\end{tabular}
\end{table*}

Panel B varies only the temporal parameters after current evidence escalation has formed $L_t$. The valley gate remains stable across nearby $H$, $\tau$, and $\eta$ values, while the one-step correction-carry coefficient $\rho_c$ produces only small changes at decision anchors; these parameters therefore affect temporal maintenance without changing PAR acquisition or SEE current-window escalation.

\paragraph{Semantic-probe limitation.}
Q3 is intentionally a dynamics probe rather than a generic anomaly classifier, and its behavior exposes a structural bias in the current factorization. Among anomalous decisions, $P(\ell_{q_3}^{(t)}<0\mid GT{=}1)$ is high for persistent categories such as MSAD Fire (66.3\%) and UCF-Crime Arson (70.6\%), but low for visibly dynamic Fighting (1.7\% on MSAD and 1.3\% on UCF-Crime). A negative Q3 both suppresses the local score and closes the P3/P4 route, so persistent environmental hazards can be penalized twice: weak dynamics evidence lowers $L_t$ and prevents the environmental specialist from verifying the condition. Category-level replay is consistent with this distinction. The present hierarchy is therefore better matched to active, interactive anomalies than to all persistent surveillance anomalies. A natural mitigation is to decouple P4 environmental routing from Q3 when coarse Q2 evidence is strong, or to treat negative dynamics evidence as neutral for non-agentic hazard categories; we leave such changes for future work because they alter the deployed operator.

\section{Limitations}
PARSEE-VAD inherits the perceptual limitations of its frozen multimodal backbone, but the current proposition hierarchy also imposes its own inductive bias. Q3 explicitly asks for active physical change and gates both specialists; consequently, persistent hazards such as already-burning fire, smoke, stationary fallen persons, or already-damaged objects can be suppressed even when Q2 remains positive. This is not solely a backbone failure: it follows in part from the prompt and routing design. The proposition readout can expose and PA-routing can allocate existing semantic competence more selectively, but neither can manufacture missing evidence, and the present hierarchy is better matched to active/interactive anomalies than to every static or diffuse surveillance condition. Decoupling environmental routing from Q3, calibrating proposition margins per backbone, targeted probe-specific adaptation, or stronger scene-level backbones are natural extensions.

The SEE current-evidence rule is deliberately low-dimensional, so errors in proposition geometry can still propagate to $L_t$. SEE's temporal memory is short-horizon and can maintain only evidence that earlier windows have already established; long evidence gaps, severe occlusions that erase coarse Q2 evidence, and anomalies absent from both current and stored scalar evidence remain unresolved. In particular, the current proposition hierarchy may require adaptation for subtle pedestrian-rule violations or out-of-distribution anomaly types that are less dominated by active physical interaction. PARSEE operates at decision anchors and therefore carries bounded decision delay. The completed-interval and anchor-aligned evaluations characterize two score-mapping views of the same causal stream, while the measured runtime characterizes the additional compute delay after each anchor and published online methods still differ in decision cadence. Runtime measurements likewise reflect the specific RTX 5880 Ada setup used here. The three deployment variants were placed on separate same-model GPUs, but host CPU/storage affinity was not isolated, so synchronized model-section timing is the cleaner compute comparison and end-to-end timing should be read as system-specific. Deployment on resource-limited hardware will require separate throughput, memory, and energy characterization.

\end{document}